\documentclass[journal]{IEEEtran}

\usepackage{framed}
\usepackage{amsmath,amssymb,amsfonts}
\usepackage{graphicx}
\usepackage{adjustbox}
\usepackage{xcolor}
\usepackage{multirow}
\usepackage{hyperref}
\def\BibTeX{{\rm B\kern-.05em{\sc i\kern-.025em b}\kern-.08em
    T\kern-.1667em\lower.7ex\hbox{E}\kern-.125emX}}
    
\usepackage{breqn}
\usepackage{algorithm}
\usepackage{algpseudocode}
\usepackage{pifont}
\usepackage{listings}
\usepackage{float}
\usepackage{rotating}
\usepackage{cite}
\usepackage{textcomp}

\def\BibTeX{{\rm B\kern-.05em{\sc i\kern-.025em b}\kern-.08em
    T\kern-.1667em\lower.7ex\hbox{E}\kern-.125emX}}
    
\begin{document}

\title{
A Framework for Neural Topic Modeling of Text Corpora
}

\author{Shayan Fazeli,~\IEEEmembership{Member,~IEEE,}
        Majid Sarrafzadeh,~\IEEEmembership{Senior Member,~IEEE}
}

\maketitle

\begin{abstract}
Topic Modeling refers to the problem of discovering the main topics that have occurred in corpora of textual data, with solutions finding crucial applications in numerous fields. In this work, inspired by the recent advancements in the Natural Language Processing domain, we introduce FAME, an open-source framework enabling an efficient mechanism of extracting and incorporating textual features and utilizing them in discovering topics and clustering text documents that are semantically similar in a corpus. These features range from traditional approaches (e.g., frequency-based) to the most recent auto-encoding embeddings from transformer-based language models such as BERT model family. To demonstrate the effectiveness of this library, we conducted experiments on the well-known News-Group dataset. The library is available online\footnote{GitHub: \href{https://github.com/shayanfazeli/fame}{https://github.com/shayanfazeli/fame}}
\end{abstract}


%
\IEEEpeerreviewmaketitle

\section{Introduction}

When it comes to analyzing a large number of documents, having to make sense of them manually is most often a cumbersome task to carry out, if not utterly impossible \cite{boyd2017applications,jelodar2019latent}. 
For many applications, users have to perform an in-depth analysis of a large number of text data and try to discover what principal points of focus can best describe the general underlying structure of the data. 
This motivates the focus on automation for performing such tasks, with an example being the similarity-based clustering of document representations, and in this way, bringing documents that are close in terms of their semantic content together. 
An application-domain instance would be going over the customer review documents and discovering what topics have often contributed to the semantic content of each document \cite{titov2008modeling,calheiros2017sentiment,sutherland2020determinants,kirilenko2021automated}.
Moreover, topic modeling and document clustering have applications in various fields, ranging from medical applications to geography and political sciences \cite{zhang2017idoctor,greene2015unveiling,jiang2012using,paul2011you,wu2012ranking,linstead2007mining,gethers2010using,asuncion2010software,thomas2011mining,cristani2008geo,eisenstein2010latent,tang2012multiscale,yin2011geographical,sizov2010geofolk,chen2010opinion,cohen2013classifying}.

Topic modeling based on Latent Dirichlet Allocation (LDA) provides the foundation for the bulk of traditional approaches to topic modeling and is still an effective approach for proposing a solution to this problem \cite{blei2003latent,jelodar2019latent}. 
Generative modeling of documents and discovering topics is possible via LDA-based pipelines. In addition to LDA-based designs, matrix factorization has also been utilized in topic modeling, and document clustering \cite{kuang2015nonnegative}.

With the advancements in employing neural network-based architectures to improve the performance in computer vision and natural language processing, the use of text embedding saw an increase in attention. From the more traditional approaches such as GloVe \cite{pennington2014glove}, the effort was focused on capturing the semantic context of each word in a real-valued vector representation such that there be a close relationship between the mathematical arithmetics and the semantic similarity of the data points. The more recent approaches focusing on auto-regressive (e.g., ElMo) and auto-encoding (e.g., Bert) models improved the state of the art performance on almost every natural language processing task by allowing the embedding of the word as well as the context it has appeared in \cite{peters2018deep,devlin2018bert}. These models, having been trained on large-scale corpora of text, are able to encode the text semantics in concise mathematical vectors effectively. It has been shown that utilizing these representations in obtaining document representations can be helpful \cite{shao_2020,green_mitchell_yu_2020}.

In this work, we propose FAME, an open-source library for designing and training neural network-based topic modeling pipelines. 
Following the aforementioned literarture, this library allows experimenting with traditional approaches to feature extraction such as LDA and term frequency–inverse document frequency (tf-idf).
It also provides the ability to utilize transformer-based document representations. It allows mixing the two domains by training auto-encoders and minimizing reconstruction loss, leading to encoded embeddings being a more concise representation associated with the different features extracted from the documents.


\section{Methods}
In what follows, the key components of FAME framework are discussed. These components help enable an efficient and easy process for designing, implementing, and evaluating the topic modeling schemes on large-scale text corpora.

\subsection{Preprocessing}
When it comes to working with large text corpora, especially for traditional approaches of topic modeling, various forms of preprocessing are needed. In addition to normal preprocessing routines (e.g., modifying or removing punctuations), it is often needed to stem and lemmatize words or perform spell-checking to fix typos in the text.
All in all, preprocessing is a critical component when it comes to deal with textual data.

\subsection{Document Representation}
\subsubsection{Term-based}
To represent a document as a real-valued vector representation, utilizing term-frequencies has been a well-known traditional approach. 
These values can either correspond directly to the term frequencies or be found by performing factorization techniques such as non-negative matrix factorization. 
In summary, tf-idf features can play an important role, especially if the dataset does not include a large number of documents.

\subsubsection{LDA-based} LDA can also be used to represent documents. 
To do so, the approach is first to fit an LDA model including a range of topics to the document corpora and then utilize it to associate each document with a probability density layout, measuring its links to the found principal topics.

\subsubsection{Transformers-based}
Transformer-based approaches focusing on Bert-family have been utilized successfully in representing documents.
Therefore, another effective way of mapping documents to a latent semantic space is employing these models. FAME employs the \texttt{sentence-transformers} library in tackling the problem of obtaining semantic word sequence representations by applying pre-trained Transformers \cite{reimers-2020-multilingual-sentence-bert,reimers-2019-sentence-bert}.

\subsection{Representation Fusion}
To be able to utilize more than one of the aforementioned feature domains, we followed the literature and provided the option to design and use an additional auto-encoder for obtaining concise representation reflecting on the information from the inputs. To train the auto-encoder,  Mean Squared Error (MSE) reconstruction loss can be used.

\subsection{Clustering}
After mapping the text documents to latent space, the similarity of the resulting representations can be effectively utilized to cluster these documents into semantically similar groups. This operation makes it easier to determine the main topics semantically similar documents have focused on.

\section{Experiments}
We conducted experiments on the 20 Newsgroups dataset to better demonstrate the results of applying the aforementioned methodologies on text corpora \cite{20newsgroup}. This dataset includes $18846$ news documents (with $7532$ of them being used as the held-out validation set), each of which is labeled with a group name from the following:

\begin{enumerate}
\item alt.atheism
\item comp.graphics
\item comp.os.ms-windows.misc
\item comp.sys.ibm.pc.hardware
\item comp.sys.mac.hardware
\item comp.windows.x
\item misc.forsale
\item rec.autos
\item rec.motorcycles
\item rec.sport.baseball
\item rec.sport.hockey
\item sci.crypt
\item sci.electronics
\item sci.med
\item sci.space
\item soc.religion.christian
\item talk.politics.guns
\item talk.politics.mideast
\item talk.politics.misc
\item talk.religion.misc
\end{enumerate}

\begin{figure}
    \centering
    \includegraphics[width=0.49\textwidth]{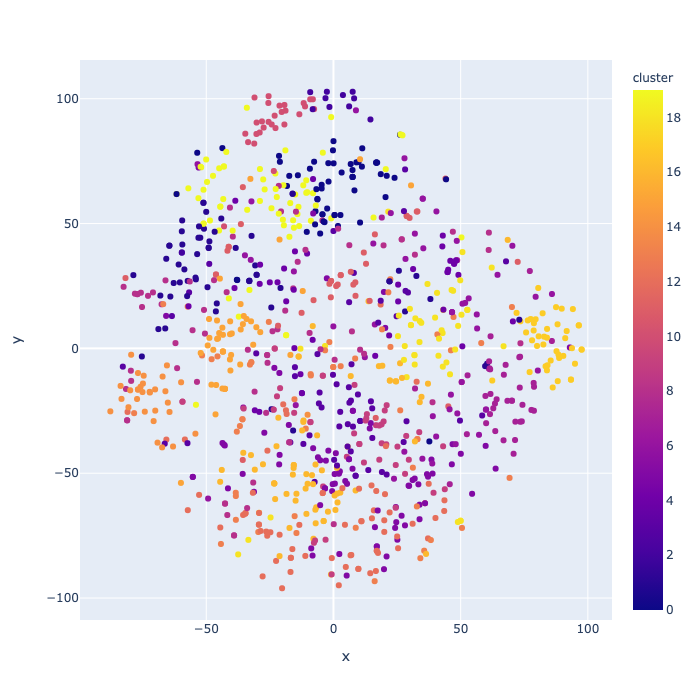}
    \caption{The 2-D t-SNE representation of document representations, with colors indicating their cluster in a 20-cluster K-Means}
    \label{fig:clustered_datapoints}
\end{figure}

\begin{figure*}
    \centering
    {\tiny 
\begin{lstlisting}
from mike avon demon us subject re distribution world organization none mike avon demon us simple news lines in article access die net access die com writes the only theory that makes any sense is that and are either the same for all chips or vary among very few possibilities so that anyone trying to break the encryption by brute force need only low through the possible serial numbers multiplied by the number of different combinations if the phones transmit their serial nos as part of the message then what is to say that each phone can take that serial number and use it to generate the required key - target: sci.crypt


from tunisia signal eye carson eds see subject re organized lobbying for cryptography tunisia signal eye carson eds see organization sun ecosystems lines signal eye carson eds in article has transfer stratus com me ellison stratus com writes to paraphrase i may not agree with what you encrypting but i defend your right to encrypt it great slogan i ready to sign up with a effort though i would want to do it through an era offshoot shall we also push for the car cryptographic rights amendment dwight tunisia best tunisia sandman eye carson eds tolerable twisted craft camp carson eds homo sapiens planetary cancer news at six - target: sci.crypt


from spectra reed eds subject re why the drive speeds differ reed organization reed college portland oregon lines in article content alana org a a content alana org a writes the most likely explanation may have something to do with the fact that a greater density of information exists on the larger capacity disk drive than the smaller one if your running the drive on a mac i would recommend a shareware utility called timeline which tests seek sci throughput and rotational speed this utility should let you know what the differences are between the drives the views expressed in this posting those of the individual author only bus number malcontent is victorious first iconic bus larger drives tend to have multiple platters which can allow adjacent bits to be read in parallel resulting in higher throughput they also have higher spindle speeds which leads to both increased throughput and reduced seek times specimen - target: comp.sys.mac.hardware


from straight sitcom com subject back doors in clipper organization lines i think it very unlikely there are back doors in clipper for two reasons the government does need them if it can get the key and yes i assume that the official government obeys court orders that the design of the chip and its approval were official it would defeat the whole purpose of providing secure crypt for american business that could be read by our economic adversaries if this were not a legitimate and genuine purpose and as many think the asa can read de why bother otherwise rational responses preferred to conspiracy theories thanks david david starlight great care has been taken to ensure the accuracy of our information errors and omissions excepted - target: sci.crypt


from access die com subject re dorothy denying opposes clipper capstone wiretap chips organization express access online communications greenbelt my us lines access die net i believe there is no technical means of ensuring key escrow without the government maintaining a secret of some kind not necessarily for instance in the system outlined in the may issue of byte the process of getting one public key listed for general use involves giving pieces of your private key to escrow agencies which do calculations on those pieces and forward the result to the publishers of the public key directory which combines these results into your listed public key if you try to give the escrow agencies pieces which yield your private key when they are all put together the result is that the public key listed for you is wrong and you a read messages encrypted to you - target: sci.crypt
\end{lstlisting}
}
    \caption{An example set of documents that have been clustered together shown with their actual group label}
    \label{fig:my_label}
\end{figure*}

\section{Conclusion}
Topic modeling and reinforcing it by integration with document representation is a critical problem with numerous applications.
Traditional and recent approaches to extract semantic features from text documents have shown to be helpful in obtaining representative clusters with semantically similar documents.
In this work, we proposed FAME, an open-source software library that enables conducting experiments related to topic modeling and document clustering with ease.

\bibliographystyle{./bibliography/IEEEtran}
\bibliography{./bibliography/IEEEabrv,./bibliography/refs.bib}

\begin{thebibliography}{10}
\providecommand{\url}[1]{#1}
\csname url@samestyle\endcsname
\providecommand{\newblock}{\relax}
\providecommand{\bibinfo}[2]{#2}
\providecommand{\BIBentrySTDinterwordspacing}{\spaceskip=0pt\relax}
\providecommand{\BIBentryALTinterwordstretchfactor}{4}
\providecommand{\BIBentryALTinterwordspacing}{\spaceskip=\fontdimen2\font plus
\BIBentryALTinterwordstretchfactor\fontdimen3\font minus
  \fontdimen4\font\relax}
\providecommand{\BIBforeignlanguage}[2]{{%
\expandafter\ifx\csname l@#1\endcsname\relax
\typeout{** WARNING: IEEEtran.bst: No hyphenation pattern has been}%
\typeout{** loaded for the language `#1'. Using the pattern for}%
\typeout{** the default language instead.}%
\else
\language=\csname l@#1\endcsname
\fi
#2}}
\providecommand{\BIBdecl}{\relax}
\BIBdecl

\bibitem{boyd2017applications}
J.~L. Boyd-Graber, Y.~Hu, D.~Mimno \emph{et~al.}, \emph{Applications of topic
  models}.\hskip 1em plus 0.5em minus 0.4em\relax Now Publishers Incorporated,
  2017, vol.~11.

\bibitem{jelodar2019latent}
H.~Jelodar, Y.~Wang, C.~Yuan, X.~Feng, X.~Jiang, Y.~Li, and L.~Zhao, ``Latent
  dirichlet allocation (lda) and topic modeling: models, applications, a
  survey,'' \emph{Multimedia Tools and Applications}, vol.~78, no.~11, pp.
  15\,169--15\,211, 2019.

\bibitem{titov2008modeling}
I.~Titov and R.~McDonald, ``Modeling online reviews with multi-grain topic
  models,'' in \emph{Proceedings of the 17th international conference on World
  Wide Web}, 2008, pp. 111--120.

\bibitem{calheiros2017sentiment}
A.~C. Calheiros, S.~Moro, and P.~Rita, ``Sentiment classification of
  consumer-generated online reviews using topic modeling,'' \emph{Journal of
  Hospitality Marketing \& Management}, vol.~26, no.~7, pp. 675--693, 2017.

\bibitem{sutherland2020determinants}
I.~Sutherland and K.~Kiatkawsin, ``Determinants of guest experience in airbnb:
  a topic modeling approach using lda,'' \emph{Sustainability}, vol.~12, no.~8,
  p. 3402, 2020.

\bibitem{kirilenko2021automated}
A.~P. Kirilenko, S.~O. Stepchenkova, and X.~Dai, ``Automated topic modeling of
  tourist reviews: does the anna karenina principle apply?'' \emph{Tourism
  Management}, vol.~83, p. 104241, 2021.

\bibitem{zhang2017idoctor}
Y.~Zhang, M.~Chen, D.~Huang, D.~Wu, and Y.~Li, ``idoctor: Personalized and
  professionalized medical recommendations based on hybrid matrix
  factorization,'' \emph{Future Generation Computer Systems}, vol.~66, pp.
  30--35, 2017.

\bibitem{greene2015unveiling}
D.~Greene and J.~P. Cross, ``Unveiling the political agenda of the european
  parliament plenary: A topical analysis,'' in \emph{Proceedings of the ACM web
  science conference}, 2015, pp. 1--10.

\bibitem{jiang2012using}
Z.~Jiang, X.~Zhou, X.~Zhang, and S.~Chen, ``Using link topic model to analyze
  traditional chinese medicine clinical symptom-herb regularities,'' in
  \emph{2012 IEEE 14th international conference on e-health networking,
  applications and services (Healthcom)}.\hskip 1em plus 0.5em minus
  0.4em\relax IEEE, 2012, pp. 15--18.

\bibitem{paul2011you}
M.~J. Paul and M.~Dredze, ``You are what you tweet: Analyzing twitter for
  public health,'' in \emph{Fifth international AAAI conference on weblogs and
  social media}, 2011.

\bibitem{wu2012ranking}
Y.~Wu, M.~Liu, W.~J. Zheng, Z.~Zhao, and H.~Xu, ``Ranking gene-drug
  relationships in biomedical literature using latent dirichlet allocation,''
  in \emph{Biocomputing 2012}.\hskip 1em plus 0.5em minus 0.4em\relax World
  Scientific, 2012, pp. 422--433.

\bibitem{linstead2007mining}
E.~Linstead, P.~Rigor, S.~Bajracharya, C.~Lopes, and P.~Baldi, ``Mining
  concepts from code with probabilistic topic models,'' in \emph{Proceedings of
  the twenty-second IEEE/ACM international conference on Automated software
  engineering}, 2007, pp. 461--464.

\bibitem{gethers2010using}
M.~Gethers and D.~Poshyvanyk, ``Using relational topic models to capture
  coupling among classes in object-oriented software systems,'' in \emph{2010
  IEEE international conference on software maintenance}.\hskip 1em plus 0.5em
  minus 0.4em\relax IEEE, 2010, pp. 1--10.

\bibitem{asuncion2010software}
H.~U. Asuncion, A.~U. Asuncion, and R.~N. Taylor, ``Software traceability with
  topic modeling,'' in \emph{2010 ACM/IEEE 32nd International Conference on
  Software Engineering}, vol.~1.\hskip 1em plus 0.5em minus 0.4em\relax IEEE,
  2010, pp. 95--104.

\bibitem{thomas2011mining}
S.~W. Thomas, ``Mining software repositories using topic models,'' in
  \emph{Proceedings of the 33rd International Conference on Software
  Engineering}, 2011, pp. 1138--1139.

\bibitem{cristani2008geo}
M.~Cristani, A.~Perina, U.~Castellani, and V.~Murino, ``Geo-located image
  analysis using latent representations,'' in \emph{2008 IEEE Conference on
  Computer Vision and Pattern Recognition}.\hskip 1em plus 0.5em minus
  0.4em\relax IEEE, 2008, pp. 1--8.

\bibitem{eisenstein2010latent}
J.~Eisenstein, B.~O’Connor, N.~A. Smith, and E.~Xing, ``A latent variable
  model for geographic lexical variation,'' in \emph{Proceedings of the 2010
  conference on empirical methods in natural language processing}, 2010, pp.
  1277--1287.

\bibitem{tang2012multiscale}
H.~Tang, L.~Shen, Y.~Qi, Y.~Chen, Y.~Shu, J.~Li, and D.~A. Clausi, ``A
  multiscale latent dirichlet allocation model for object-oriented clustering
  of vhr panchromatic satellite images,'' \emph{IEEE Transactions on Geoscience
  and Remote Sensing}, vol.~51, no.~3, pp. 1680--1692, 2012.

\bibitem{yin2011geographical}
Z.~Yin, L.~Cao, J.~Han, C.~Zhai, and T.~Huang, ``Geographical topic discovery
  and comparison,'' in \emph{Proceedings of the 20th international conference
  on World wide web}, 2011, pp. 247--256.

\bibitem{sizov2010geofolk}
S.~Sizov, ``Geofolk: latent spatial semantics in web 2.0 social media,'' in
  \emph{Proceedings of the third ACM international conference on Web search and
  data mining}, 2010, pp. 281--290.

\bibitem{chen2010opinion}
B.~Chen, L.~Zhu, D.~Kifer, and D.~Lee, ``What is an opinion about? exploring
  political standpoints using opinion scoring model,'' in \emph{Twenty-Fourth
  AAAI Conference on Artificial Intelligence}, 2010.

\bibitem{cohen2013classifying}
R.~Cohen and D.~Ruths, ``Classifying political orientation on twitter: It’s
  not easy!'' in \emph{Proceedings of the International AAAI Conference on Web
  and Social Media}, vol.~7, no.~1, 2013.

\bibitem{blei2003latent}
D.~M. Blei, A.~Y. Ng, and M.~I. Jordan, ``Latent dirichlet allocation,''
  \emph{the Journal of machine Learning research}, vol.~3, pp. 993--1022, 2003.

\bibitem{kuang2015nonnegative}
D.~Kuang, J.~Choo, and H.~Park, ``Nonnegative matrix factorization for
  interactive topic modeling and document clustering,'' in \emph{Partitional
  Clustering Algorithms}.\hskip 1em plus 0.5em minus 0.4em\relax Springer,
  2015, pp. 215--243.

\bibitem{pennington2014glove}
J.~Pennington, R.~Socher, and C.~D. Manning, ``Glove: Global vectors for word
  representation,'' in \emph{Proceedings of the 2014 conference on empirical
  methods in natural language processing (EMNLP)}, 2014, pp. 1532--1543.

\bibitem{peters2018deep}
M.~E. Peters, M.~Neumann, M.~Iyyer, M.~Gardner, C.~Clark, K.~Lee, and
  L.~Zettlemoyer, ``Deep contextualized word representations,'' \emph{arXiv
  preprint arXiv:1802.05365}, 2018.

\bibitem{devlin2018bert}
J.~Devlin, M.-W. Chang, K.~Lee, and K.~Toutanova, ``Bert: Pre-training of deep
  bidirectional transformers for language understanding,'' \emph{arXiv preprint
  arXiv:1810.04805}, 2018.

\bibitem{shao_2020}
\BIBentryALTinterwordspacing
S.~Shao, ``Contextual topic identification,'' 2020. [Online]. Available:
  \url{https://bit.ly/3zaS8k6}
\BIBentrySTDinterwordspacing

\bibitem{green_mitchell_yu_2020}
\BIBentryALTinterwordspacing
W.~Green \emph{et~al.}, ``Topic modeling bert+lda,'' 2020. [Online]. Available:
  \url{https://www.kaggle.com/dskswu/topic-modeling-bert-lda}
\BIBentrySTDinterwordspacing

\bibitem{reimers-2020-multilingual-sentence-bert}
\BIBentryALTinterwordspacing
N.~Reimers and I.~Gurevych, ``Making monolingual sentence embeddings
  multilingual using knowledge distillation,'' in \emph{Proceedings of the 2020
  Conference on Empirical Methods in Natural Language Processing}.\hskip 1em
  plus 0.5em minus 0.4em\relax Association for Computational Linguistics, 11
  2020. [Online]. Available: \url{https://arxiv.org/abs/2004.09813}
\BIBentrySTDinterwordspacing

\bibitem{reimers-2019-sentence-bert}
\BIBentryALTinterwordspacing
------, ``Sentence-bert: Sentence embeddings using siamese bert-networks,'' in
  \emph{Proceedings of the 2019 Conference on Empirical Methods in Natural
  Language Processing}.\hskip 1em plus 0.5em minus 0.4em\relax Association for
  Computational Linguistics, 11 2019. [Online]. Available:
  \url{https://arxiv.org/abs/1908.10084}
\BIBentrySTDinterwordspacing

\bibitem{20newsgroup}
\BIBentryALTinterwordspacing
2021. [Online]. Available: \url{http://qwone.com/~jason/20Newsgroups/}
\BIBentrySTDinterwordspacing

\end{thebibliography}

\end{document}